\documentclass[10pt,twocolumn,letterpaper]{article}

\usepackage[pagenumbers]{cvpr} 

\usepackage{microtype}

\renewcommand{\paragraph}[1]{\vspace{.5em}\noindent\textbf{#1.}}

\usepackage{multirow}

\usepackage{amsmath,amsfonts,bm}
\usepackage{xspace}

\def\eqref#1{equation~\ref{#1}}

\def\1{\bm{1}}

\def\vo{{\bm{o}}}

\def\vs{{\bm{s}}}

\def\vu{{\bm{u}}}
\def\vv{{\bm{v}}}

\def\vx{{\bm{x}}}

\def\vz{{\bm{z}}}

\DeclareMathAlphabet{\mathsfit}{\encodingdefault}{\sfdefault}{m}{sl}
\SetMathAlphabet{\mathsfit}{bold}{\encodingdefault}{\sfdefault}{bx}{n}

\makeatletter
\DeclareRobustCommand\onedot{\futurelet\@let@token\@onedot}
\def\@onedot{\ifx\@let@token.\else.\null\fi\xspace}

\def\eg{\emph{e.g}\onedot}

 \def\vs{\emph{vs}\onedot}

\makeatother

\definecolor{cvprblue}{rgb}{0.21,0.49,0.74}
\usepackage[pagebackref,breaklinks,colorlinks,allcolors=cvprblue]{hyperref}

\def\paperID{16} 
\def\confName{CVPR}
\def\confYear{2026}

\title{Beyond Flat Labels: Level-Restricted Contrastive Learning for\\ Hierarchical Fine-Grained Vision Classification}

\author{Zhiyuan Tao$^{1\dagger}$\quad Srikumar Sastry$^2$\quad Matthew J Thompson$^1$\quad Elizabeth G Campolongo$^1$\\ Net Zhang$^1$\quad Ziheng Zhang$^1$ \quad Hilmar Lapp$^3$\quad Yu Su$^1$\quad Tanya Berger-Wolf$^1$\\ Nathan Jacobs$^2$\quad Wei-Lun Chao$^4$\quad Jianyang Gu$^{1\dagger}$\\
{\small $^1$ The Ohio State University\quad $^2$ Washington University in St. Louis\quad $^3$ Neuromatch\quad $^4$ Boston University}\\
$^\dagger${\tt\small \{tao.623, gu.1220\}@osu.edu}
}

\begin{document}
\maketitle
\begin{abstract}
Multimodal contrastive learning has enabled zero-shot visual classification by aligning images with textual categories. However, in hierarchically structured label spaces, existing methods often produce predictions that are inconsistent across taxonomic levels. For example, a model may predict a fine-grained category whose parent category contradicts its simultaneously predicted higher-level label. By analysis, the issue originates from false negative labels when contrastive comparison involves multiple taxonomic levels. To this end, we propose to restrict contrastive comparisons to categories within the same taxonomic level. In addition, we adopt a group-balanced design, ensuring each taxonomic level receives adequate optimization. As a result, the proposed framework improves both hierarchical consistency and classification accuracy from coarse to fine granularity. We train our model with TreeOfLife-10M based on BioCLIP and evaluate it across multiple hierarchical classification benchmarks, where the model demonstrates significantly improved hierarchical consistency in both Euclidean and hyperbolic spaces. Notably, on iNaturalist 2021 (iNat21), our method improves average accuracy across levels by 30.47\% over the baseline, highlighting its effectiveness for hierarchical zero-shot classification. We have released our \href{https://huggingface.co/imageomics/bioclip-hc-euclidean}{Euclidean} and \href{https://huggingface.co/imageomics/bioclip-hc-hyperbolic}{hyperbolic} models on HuggingFace. 
\end{abstract}
    
\section{Introduction}
\label{sec:intro}

Contrastive learning between images and text has produced strong models capable of zero-shot fine-grained visual classification, such as CLIP~\cite{radford2021learning}, ALIGN~\cite{jia2021align}, and CoCa~\cite{yu2022coca}. However, these models often suffer from hierarchical inconsistency when predicting categories across taxonomic levels. That is, the predicted fine-grained label does not belong to the predicted higher-level category. Such inconsistency leads to ambiguity for real-world applications~\cite{park2025visually}.

\begin{figure*}[t]
  \centering
  \includegraphics[width=0.8\linewidth, trim=0.5cm 4.8cm 0cm 3.5cm, clip]{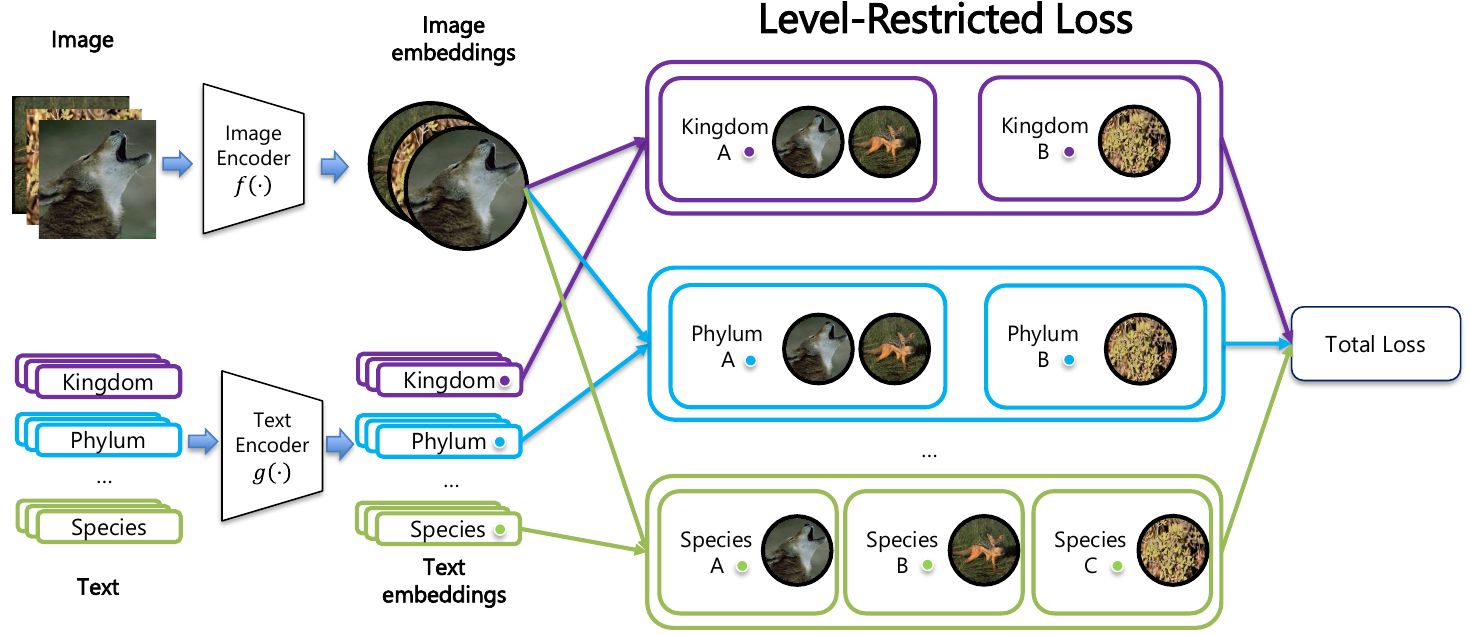}
  \vspace{-4pt}
  \caption{\textbf{Level-restricted contrastive learning.}
  For each image, we construct taxonomic text labels at all hierarchical levels and encode images and texts using CLIP encoders. Instead of comparing text labels from all levels jointly, we constrain the contrastive comparison to be \emph{within the same level}. This removes cross-level false negatives and enables balanced supervision across hierarchy levels.}
  \label{fig:method}
\end{figure*}

Common strategies to improve hierarchical consistency include additionally-defined geometry objectives~\cite{sastry2025rcme, alper2024radial,nickel2017poincare,nickel2018lorentz,desai2023meru} or hierarchy-oriented architectures~\cite{park2025visually,wehrmann2018hmcn,chen2022hrn,zhu2017bcnn}.
In this work, we focus on the contrastive training objective to understand the gap from hierarchical consistency.
Standard CLIP-style contrastive learning adopts a flat-label assumption and treats all non-matching text labels as equal negatives~\cite{radford2021learning}. 
When applied to label spaces with hierarchical relationships, however, such flat comparison might lead to false negatives.
While treating ancestor--descendant labels as negatives is clearly incorrect, this issue is not limited to direct parent--child relationships.
Consider images of \textit{Panthera leo} (lion) and \textit{Canis lupus} (wolf). They are distinct at the species and genus levels, but they share higher-level ancestors such as the order \textit{Carnivora}, which makes them semantically compatible at coarser levels.
As a result, when labels from multiple taxonomic levels are combined within a single contrastive objective, there is no consistent way to define negative pairs, since samples that should be separated at one level may be similar at another. This violates the flat-label assumption underlying contrastive learning, leading to conflicting supervision signals.

To address this issue, we restrict contrastive comparisons to operate within a single taxonomy level, where labels are mutually exclusive, and the contrastive formulation remains well-defined. For example, a species-level label is compared only with other species-level labels. Such a restriction avoids conflicts or ambiguity when comparing labels from different hierarchical levels. In addition, we introduce group-balanced supervision across levels to ensure that each level receives adequate optimization, so that training will not be biased toward coarse hierarchical levels.

We conduct experiments with biodiversity data, based on the previous foundation models on the tree of life~\cite{yang2024biotrove, stevens2024bioclip, sastry2025taxabind,gu2025bioclip}. Specifically, we finetune BioCLIP~\cite{stevens2024bioclip,bioclip2023} with TreeOfLife-10M~\cite{stevens2024bioclip,treeoflife_10m}, following the proposed contrastive recipe.
The derived model is evaluated across three species classification benchmarks (iNat21~\cite{inat2021}, Rare Species~\cite{rare_species_2023,stevens2024bioclip}, and CrypticBio~\cite{manolache25crypticbio}), demonstrating more consistent hierarchical predictions. On iNat21~\cite{inat2021}, our model improves the average accuracy by more than 30\% in both Euclidean and hyperbolic embedding spaces.
Qualitatively, we show that our model learns text embeddings that form hierarchical structures. 

\section{Method}
\label{sec:method}

Figure~\ref{fig:method} illustrates the proposed training framework.
Given an image, we construct text labels for all taxonomic levels
(kingdom $\rightarrow$ species) and encode images and text using corresponding CLIP encoders.
Rather than performing contrastive learning over labels from all levels jointly,
we reformulate the objective to operate in a \emph{level-restricted} manner.

\subsection{Problem Setup}

We consider a taxonomy set $\mathcal{T}=(\mathcal{C}^{1},\ldots,\mathcal{C}^{L})$ with $L$ hierarchical levels from coarse to fine, where $\mathcal{C}^{\ell}=\{t^{\ell}\}$ contains all the labels belonging to level $\ell$. Each image $\vx_i$ is associated with a taxonomic label $t_i=\left(t_i^{1},\ldots,t_i^{L}\right)$, where $t_i^{\ell} \in \mathcal{C}^{\ell}$.
We use the CLIP image encoder $f(\cdot)$ to produce a unique image embedding $\vv_i$ for the sample.
On the text side, we first construct multi-level taxonomic text labels:
\[
\mathcal{Y}_i=\{y_i^1,\ldots,y_i^L\},\quad y_i^\ell=(t^1_i,\ldots,t^\ell_i).
\]
Each level’s text label contains the taxonomy prefix up to that level.
Then we use the text encoder $g(\cdot)$ to extract text embeddings $\vz_i^{\ell}=g(y_i^\ell)$ at level $\ell$.

\begin{table*}[t]
\centering
\small
{
\begin{tabular}{ll|ccccccc|c}
\toprule
Method & Space & Kingdom 
& Phylum & Class & Order & Family & Genus & Species & Avg. \\
\midrule

OpenCLIP & Euclidean  
& 84.76 & 35.37 & 26.08 & 19.25 & 7.71 & 6.80 & 2.09 & 26.01 \\

BioCLIP & Euclidean  
& 86.28 & 56.14 & 41.69 & 26.95 & 30.37 & 47.21 & 50.79 & 48.49 \\

\midrule
RCME & Euclidean
& 86.26 & 83.00 & 70.79 & 46.46 & 44.74 & 59.28 & 50.50 & 63.00 \\

\midrule
\multirow{2}{*}{Ours} 
& Euclidean 
& 98.85 & 98.39 & 67.23 & \textbf{78.22} & 78.69 & 68.36 & \textbf{63.00} & \textbf{78.96} \\

& Hyperbolic 
& \textbf{98.97} & \textbf{98.48} & \textbf{71.81} & 75.84 & \textbf{82.25} & \textbf{73.96} & 51.35 & 78.95 \\

\bottomrule
\end{tabular}
}
\vspace{-2pt}
\caption{Per-level and average top-1 accuracy (\%) on iNat21 across all taxonomic levels}
\label{tab:iNat21_results}
\end{table*}

\begin{table*}[t]
\centering
\small
{
\begin{tabular}{ll|ccccccc|c}
\toprule
Method & Space 
& Kingdom & Phylum & Class & Order & Family & Genus & Species & LCA \\
\midrule

OpenCLIP & Euclidean 
& 84.76 & 30.44 & 10.62 & 3.81 & 0.78 & 0.28 & 0.12 & 0.19 \\

BioCLIP & Euclidean 
& 75.38 & 61.34 & 48.22 & 16.85 & 9.71 & 6.73 & 4.80 & 0.32 \\

\midrule
RCME & Euclidean 
& 86.27 & 78.46 & 63.97 & 29.59 & 18.25 & 13.00 & 8.92 & 0.43 \\

\midrule
\multirow{2}{*}{Ours}
& Euclidean 
& 98.85 & 98.41 & 74.52 & 68.28 & 60.31 & 47.92 & 33.64 & 0.69 \\

& Hyperbolic 
& \textbf{98.97} & \textbf{98.53} & \textbf{77.15} & \textbf{69.03} & \textbf{62.79} & \textbf{52.64} & \textbf{36.89} & \textbf{0.71} \\

\bottomrule
\end{tabular}
}
\vspace{-2pt}
\caption{Per-level accuracy (\%) and normalized LCA under top-down constrained inference on iNat21. Predictions proceed from coarse to fine with candidate labels restricted to valid descendants, enforcing hierarchical consistency. Normalized LCA measures how deep predictions remain correct along the taxonomy.}
\label{tab:topdown_inat}
\vspace{-6pt}
\end{table*}

\subsection{Level-Restricted Contrastive Learning}

Standard contrastive learning assumes a flat, mutually exclusive label space, with only an image and its text label forming a positive pair. The other text labels are treated as equal negative targets. 
In a hierarchical classification problem, however, categories belonging to the same higher-level taxonomy share common features (\eg, lion and wolf belong to the same order), yet standard CLIP treats them as negatives equal to categories from other taxonomy paths (\eg, plants). Such properties lead to conflicting supervision signals across taxonomic levels. Inspired by this, we propose to decompose training into independent multi-level comparisons. More specifically, we calculate contrastive loss for each taxonomic level, and each loss will only involve labels of its corresponding level. 

However, a mini-batch may inevitably contain samples with the same taxonomic label, especially for coarse levels. For instance, images of two species are from the same kingdom. Only using an image and its corresponding label to form positive pairs leads to conflicting supervision---the same label is treated as both positive and negative targets. Therefore, we propose to aggregate the labels within a mini-batch and treat all image–text pairs sharing the same label as positives. 
To simplify notation, throughout the following level-wise definitions, we omit the dependence on $\ell$ when no ambiguity arises. We use $\vz_i$ to represent the text embedding of $\vx_i$ at the specific level $\ell$.
We define $\mathcal{K}=\{\vz_k\}_{k=1}^{K\le N}$ to be the text embedding set of the corresponding level, where $N$ is the mini-batch size.
Accordingly, the image-to-text loss can be calculated as:
\begin{equation}
\label{eq:i2t}
\mathcal{L}^{I\rightarrow T}
=
-\frac{1}{N}\sum_{i=1}^{N}
\log
\frac{\exp(\langle\vv_i,\vz_i\rangle/\tau)}
{\sum_{\vz_k \in \mathcal{K}} \exp(\langle\vv_i,\vz_k\rangle/\tau)}.
\end{equation}
Thereby, similarity is computed between images and the set of unique text labels, avoiding conflicting supervision. 

On the other hand, each text label has multiple positive image targets. Therefore, we define a positive image set $\mathcal{P}_k=\{\vv_p\}$ for each text anchor $\vz_k$. We calculate the text-to-image loss by:
\begin{equation}
\label{eq:t2i}
\mathcal{L}^{T\rightarrow I}
=
-\frac{1}{|\mathcal{K}|}
\sum_{\vz_k \in \mathcal{K}}
\frac{1}{|\mathcal{P}_k|}
\sum_{\vv_p \in \mathcal{P}_k}
\log
\frac{\exp(\left<\vv_p,\vz_k\right>/\tau)}
{\sum_{j=1}^{N}\exp(\left<\vv_j,\vz_k\right>/\tau)}.
\end{equation}
This group-balanced strategy also mitigates category bias within each mini-batch (some categories may appear more often than others).
While we instantiate the similarity in Euclidean space, it can also be extended to hyperbolic space. We provide more details in the supplementary material.

The objective $\mathcal{L}_\ell$ at level $\ell$ is the sum of $\mathcal{L}^{I\rightarrow T}_{\ell}$ and $\mathcal{L}^{T\rightarrow I}_{\ell}$.
The final training objective is obtained by:
\begin{equation}
\mathcal{L}
=
\frac{1}{L}\sum_{\ell=1}^{L}\mathcal{L}_{\ell}.
\end{equation}
Assigning equal weight to each hierarchy level ensures sufficient supervision for fine-grained distinctions, avoiding domination by high-confidence coarse-level signals.

\begin{table*}[t]
\centering
\small
{
\begin{tabular}{ll|cccccc|c}
\toprule
Method & Space 
& Phylum & Class & Order & Family & Genus & Species & Avg. \\
\midrule

OpenCLIP & Euclidean 
& 75.89 & 60.87 & 33.32 & 13.31 & 15.27 & 10.62 & 34.88 \\

BioCLIP & Euclidean  
& 68.35 & 65.16 & 54.63 & 40.01 & 47.43 & 31.82 & 51.23 \\

\midrule
RCME & Euclidean 
& 82.38 & 81.67 & 69.01 & 44.12 & 50.95 & 35.58 & 60.62 \\

\midrule
\multirow{2}{*}{Ours} 
& Euclidean 
& \textbf{98.19} & \textbf{93.02} & 83.27 & 67.41 & 55.08 & 40.68 & 72.94 \\

& Hyperbolic
& 97.99 & 92.64 & \textbf{83.33} & \textbf{69.06} & \textbf{57.22} & \textbf{43.20} & \textbf{73.91} \\

\bottomrule
\end{tabular}
}
\vspace{-4pt}
\caption{Per-level and average top-1 accuracy (\%) on RareSpecies across all taxonomic levels}
\label{tab:RareSpecies_results}
\vspace{-8pt}
\end{table*}

\begin{table}[t]
\centering
\small
\resizebox{\columnwidth}{!}{
\begin{tabular}{ll|cccc|c}
\toprule
Method & Space 
& Order & Family & Genus & Species & Avg. \\
\midrule

OpenCLIP & Euclidean 
& 55.37 & 15.92 & 6.41 & 6.17 & 20.97 \\

BioCLIP & Euclidean 
& 81.34 & 39.29 & 37.71 & 36.72 & 48.77 \\

\midrule
RCME & Euclidean 
& 90.11 & 49.84 & 39.45 & 38.97 & 54.59 \\

\midrule
\multirow{2}{*}{Ours}
& Euclidean 
& \textbf{97.98} & 72.65 & 46.30 & 40.37 & 64.33 \\

& Hyperbolic 
& 96.80 & \textbf{78.01} & \textbf{52.18} & \textbf{49.98} & \textbf{69.24} \\

\bottomrule
\end{tabular}
}
\vspace{-4pt}
\caption{Per-level and average top-1 accuracy (\%) on CrypticBio across all taxonomic levels}
\label{tab:CrypticBio_results}
\vspace{-8pt}
\end{table}

\section{Experiments}

\subsection{Experimental Setup}
\label{sec:exp:setup}

We fine-tune BioCLIP~\cite{stevens2024bioclip,bioclip2023} on the original training set TreeOfLife-10M~\cite{treeoflife_10m,stevens2024bioclip} and evaluate on three benchmarks: iNat21~\cite{inat2021}, which provides broad taxonomic coverage; Rare Species~\cite{rare_species_2023,stevens2024bioclip}, which focuses on rare and long-tailed species absent from TreeOfLife-10M; and CrypticBio~\cite{manolache25crypticbio}, which emphasizes visually ambiguous species.

\paragraph{Metrics}
We report per-level top-1 accuracy and the average accuracy across all taxonomic levels. 
To evaluate hierarchical consistency, we adopt a \emph{top-down constrained inference} protocol~\cite{silla2011survey}, where predictions are performed sequentially from coarse to fine levels, and the candidate label set at each level is restricted to the valid descendants of the predicted parent category. Therefore, for certain levels, the derived accuracy might be higher as the candidate set is smaller. Conversely, wrong predictions at the parent levels lead to lower accuracy at lower levels. We additionally report normalized Lowest Common Ancestor (nLCA) depth~\cite{kosmopoulos2015lca}, which measures the depth of agreement between the predicted and ground-truth taxonomic paths, normalized by the total hierarchy depth~$L$.

\subsection{Main Results}
\label{sec:sota_comparison}

Tables~\ref{tab:iNat21_results},~\ref{tab:RareSpecies_results}, and~\ref{tab:CrypticBio_results} compare our method with OpenCLIP~\cite{ilharco_gabriel_2021_5143773_openclip}, BioCLIP~\cite{stevens2024bioclip,bioclip2023}, and RCME~\cite{sastry2025rcme} on the three benchmarks, respectively. Our method achieves the best average accuracy on all settings. Compared with RCME, which also fine-tunes BioCLIP with TreeOfLife-10M, our best variant improves average accuracy by more than 13\% across all benchmarks. These gains are consistent across datasets with broad taxonomic coverage, species absent in training, and visually ambiguous fine-grained categories.

The improvement is achieved across the hierarchy, and is especially prominent at intermediate levels (order–genus), where standard contrastive loss suffers most from cross-level ambiguity. Although the optimization is not solely focused on the species level, the group-balance design and reduced conflicting supervision signals also improve the species-level performance. This suggests that level-restricted contrastive supervision preserves hierarchical structure while also improving fine-grained discrimination.

The advantage of our method becomes even clearer under top-down classification (Table~\ref{tab:topdown_inat}), a stricter evaluation protocol in which an error at a higher taxonomic level constrains all subsequent predictions. Despite this compounding effect, our method remains substantially stronger than all baselines across all levels. In particular, the hyperbolic variant achieves the best nLCA score (0.71 \vs 0.43 for RCME), indicating that its predictions remain closer to the correct taxonomic branch even when the exact label is incorrect. The gains are especially large at deeper levels, improving accuracy over RCME by 44.54\%, 39.64\%, and 27.97\% at family, genus, and species levels, respectively. These results show that our method not only improves flat-recognition accuracy but also yields more hierarchically consistent predictions.

The Euclidean and hyperbolic variants show complementary behavior. The Euclidean variant performs better on iNat21 and achieves the highest species accuracy there, while the hyperbolic variant performs better on Rare Species and CrypticBio, particularly at deeper taxonomic levels. In the top-down setting, hyperbolic space demonstrates a consistent advantage over Euclidean space. Overall, these results suggest that the proposed hierarchical contrastive objective improves hierarchical consistency in both spaces, while hyperbolic space yields a stronger capability in representing hierarchical label structures.

\begin{figure}[!t]
\centering
\includegraphics[
  width=0.46\textwidth,
  trim=7cm 2cm 10cm 2cm,  
  clip
]{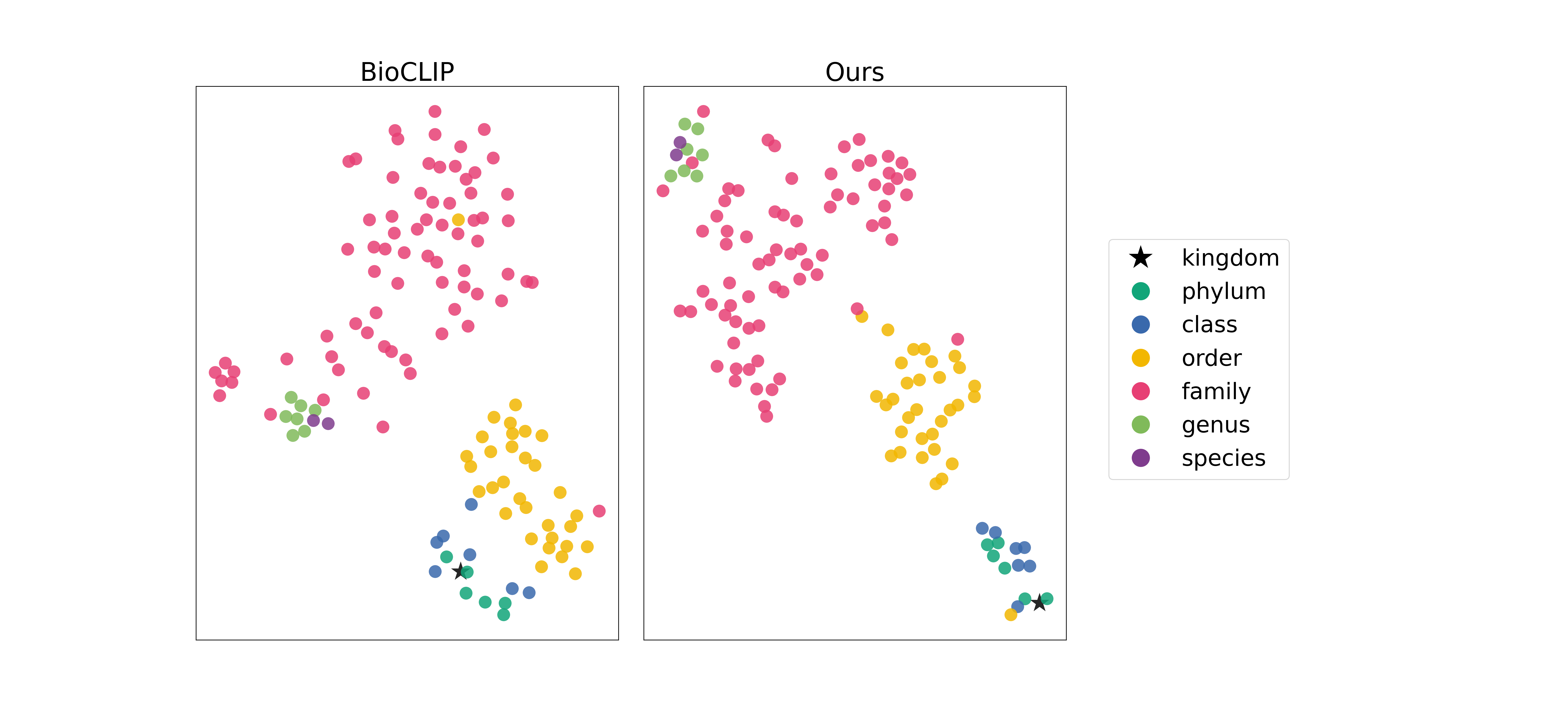}
\vspace{-8pt}
\caption{t-SNE visualization of text embeddings. Our method shows a clearer hierarchical structure and reduced cross-level mixing compared with BioCLIP.}
\label{fig:bioclip_vs_ours}
\vspace{-12pt}
\end{figure}

\subsection{Representation Visualization} 
\label{sec:repr_viz} 
We follow the same visualization protocol as~\cite{sastry2025rcme}. Specifically, we select a target species and collect its labels across all taxonomic levels. For each level $\ell$, we consider samples whose higher-level taxonomy matches that of the target species, and visualize the corresponding text embeddings for the target label and its sibling categories.
Figure~\ref{fig:bioclip_vs_ours} shows that our method produces a much clearer hierarchical structure of the text embedding space than BioCLIP. In our model, embeddings from nearby taxonomic levels form more coherent and structured groups, with reduced cross-level mixing and clearer separation among sibling categories. By contrast, BioCLIP exhibits a more scattered arrangement, where relationships across taxonomic levels are less organized. This qualitative pattern is consistent with the improvements in hierarchical evaluation metrics such as nLCA, suggesting that our method better preserves taxonomic structure in the embedding space.
\section{Conclusion}
This work studies multimodal contrastive learning for hierarchical classification. When standard contrastive learning is applied to hierarchical labels, semantically compatible ancestor--descendant concepts can be treated as false negatives. Such conflicting supervision signals lead to hierarchical inconsistency. We propose a level-restricted contrastive training framework to perform contrastive comparison within each taxonomic level, with group-balanced supervision that balances the focus on each level. Experiments show consistent gains in both classification accuracy and hierarchical consistency. 
Qualitatively, we show that level-restricted contrastive learning promotes the emergence of hierarchical structures in the embedding space.

\paragraph{Acknowledgment}
This work was supported by the U.S. National
Science Foundation (OAC-2118240) and resources from the Ohio Supercomputer Center~\cite{OhioSupercomputerCenter1987}.

{
    \small
    \bibliographystyle{ieeenat_fullname}
    \bibliography{main}
}

\clearpage
\setcounter{page}{1}
\maketitlesupplementary

\section{Related Work}
\label{sec:related-work}

\subsection{Hierarchical Image Classification}

Hierarchical image classification incorporates taxonomy structure into visual recognition~\cite{silla2011survey}. Early approaches introduced hierarchical supervision into convolutional networks to model coarse-to-fine label dependencies~\cite{zhu2017bcnn, wehrmann2018hmcn}. More recent methods integrate hierarchy into deep architectures through feature sharing, taxonomy-aware prediction, or cross-level interaction mechanisms~\cite{chen2022hrn,jiang2024ps-bkt,xia2025hgclip}.

Another line of work focuses on hierarchy-aware objectives, where prediction errors are weighted according to their distance in the taxonomy tree~\cite{bertinetto2020making}. Hierarchical supervision has also been explored in contrastive learning, for example, through losses that incorporate label similarity~\cite{lian2024lascl} or multi-level contrastive learning strategies~\cite{ghanooni2025mlcl}.

Rather than introducing specialized architectures or explicit hierarchy-aware penalties, this work focuses on how standard contrastive supervision itself becomes mismatched under hierarchical labels and redesigns the objective to better align with the taxonomy structure.

\subsection{Hierarchical Representation Learning}

A common approach to hierarchical representation learning is to impose structure through geometry-aware embedding spaces. Hyperbolic representations are particularly effective for tree-like data because of their exponential volume growth~\cite{bridson2013metric}. This idea has been explored in a range of hierarchical embedding methods, including Poincaré embeddings~\cite{nickel2017poincare}, Lorentz embeddings~\cite{nickel2018lorentz}, and more recent multimodal models such as MERU~\cite{desai2023meru}, ATMG~\cite{ramasinghe2024atmg}, and radial hyperbolic learning approaches~\cite{alper2024radial}.

Beyond geometry alone, recent work has incorporated explicit hierarchical regularization or compositional structure into representation learning, for example, through hyperbolic hierarchical regularization~\cite{sinha2024hypstructure} and compositional vision--language models that encode hierarchical relationships between images, regions, and text~\cite{pal2024hycoclip}.

This work focuses on the contrastive objective itself. We revisit hierarchical learning from an objective-design perspective, asking whether hierarchical structures can emerge in the embedding space without explicit guidance.

\subsection{CLIP for Biodiversity}

Contrastive language--image pretraining methods such as CLIP~\cite{radford2021learning}, ALIGN~\cite{jia2021align}, and CoCa~\cite{yu2022coca} have established a strong foundation for visual recognition by learning aligned representations from large-scale image--text pairs. Recent efforts have extended this paradigm to biodiversity recognition. For example, BioCLIP~\cite{stevens2024bioclip} trains a vision foundation model on TreeOfLife-10M, while BioTrove~\cite{yang2024biotrove} and TaxaBind~\cite{sastry2025taxabind} further explore large-scale image--text learning for taxonomic recognition. BioCAP leverages synthetic captions to introduce supplementary information beyond species labels~\cite{zhang2026biocap}.

Despite strong species-level performance, these models are typically trained with contrastive objectives designed for flat label spaces. When applied to hierarchical taxonomies, such objectives can create supervision mismatch across levels, leading to predictions that are inconsistent across ranks. For example, when the model predicts both species and genus labels, the species might not belong to the genus. This hierarchical inconsistency motivates methods that more effectively preserve the hierarchical structure.

\section{Implementation Details}

We initialize from the BioCLIP ViT-B/16 model~\cite{bioclip2023,stevens2024bioclip} and fine-tune it for 30 epochs using AdamW with batch size 4096, learning rate $10^{-4}$, and weight decay 0.2. A single shared temperature parameter is used across all taxonomy levels. Taxonomic text labels are constructed using cumulative taxonomy prefixes; for example, a species-level prompt includes all ancestor ranks along the taxonomic path. For hyperbolic experiments, we use fixed curvature $c=1$.

\paragraph{Hyperbolic embedding and similarity}
We adopt the Lorentz (hyperboloid) model of hyperbolic space~\cite{nickel2018lorentz} with fixed curvature $c=1$. For vectors $\vu,\vv \in \mathbb{R}^{d+1}$, the Minkowski inner product is defined as
\[
\langle \vu, \vv \rangle_{\mathcal{L}} = -u_0 v_0 + \sum_{j=1}^{d} u_j v_j ,
\]
and the hyperboloid is
\[
\mathbb{H}^{d}=\{\vx \in \mathbb{R}^{d+1} \mid \langle \vx,\vx\rangle_{\mathcal{L}}=-1/c,\ x_0>0\}.
\]
Encoder outputs are \emph{not} $\ell_2$-normalized. Following MERU~\cite{desai2023meru}, we scale them by learnable scalars $e^{\alpha_v}$ (image) and $e^{\alpha_t}$ (text) and lift them onto the manifold with the exponential map at the origin $\vo$:
\[
\vv_i = \exp_{\vo}\!\big(e^{\alpha_v} f(\vx_i)\big), \qquad
\vz_k^{(\ell)} = \exp_{\vo}\!\big(e^{\alpha_t} g(y_k^{\ell})\big).
\]
We measure image--text similarity with the oxy-angle~\cite{desai2023meru}, an angle-based quantity defined on the hyperbolic triangle formed by the origin and the two embeddings, rather than the Lorentz geodesic distance. The similarity is the negative oxy-angle divided by the temperature $\tau$, and it replaces $\langle\cdot,\cdot\rangle$ in the image-to-text and text-to-image losses (equations~\ref{eq:i2t} and~\ref{eq:t2i} in the main paper). The same similarity is used for zero-shot inference.

\end{document}